\documentclass[pmlr,twocolumn,10pt]{jmlr} %

\usepackage{booktabs}
\usepackage{siunitx}

\usepackage{float}

\usepackage{adjustbox}

\usepackage[normalem]{ulem}
\useunder{\uline}{\ul}{}

\newcommand{\Iendo}{I_{\text{endo}}}

\definecolor{myred}{rgb}{1.0, 0.3882, 0.2784}
\definecolor{mygreen}{rgb}{0.20784314, 0.67843137, 0.6}
\newcommand{\yes}{\textcolor{mygreen}{Yes}}
\newcommand{\no}{\textcolor{myred}{No}}

\theorembodyfont{\upshape}
\theoremheaderfont{\scshape}
\theorempostheader{:}
\theoremsep{\newline}

\jmlrvolume{225}
\jmlryear{2023}
\jmlrsubmitted{LEAVE UNSET}
\jmlrpublished{LEAVE UNSET}
\jmlrworkshop{Machine Learning for Health (ML4H) 2023} %

\title[Interpretable Mechanistic Representations for Glycemic Control]{
Interpretable Mechanistic Representations for Meal-level Glycemic Control in the Wild
}

\usepackage[format=plain]{caption}  %
 \usepackage{lipsum}

\newcommand*{\cbf}{\mathbf{c}}

\newcommand*{\ubf}{\mathbf{u}}

\newcommand*{\wbf}{\mathbf{w}}
\newcommand*{\xbf}{\mathbf{x}}

\newcommand*{\zbf}{\mathbf{z}}

\newcommand*{\Lcal}{\mathcal{L}}

\newcommand\given[1][]{\:#1\vert\:}
\newcommand*{\ev}[2][]{\mathbb{E}_{#1}\left[ #2\right]}
\newcommand*{\pdiv}[3][]{\operatorname{D}_{#1}\left(#2\:\vert\vert\:#3 \right)}
\newcommand*{\kldiv}[2]{\pdiv[\text{KL}]{#1}{#2}}
\newcommand*{\reals}{\mathbb{R}}

\newcommand{\reveal}[1]{#1}
 \DeclareUnicodeCharacter{03B2}{\ensuremath{\beta}} %

\author{%
\Name{Ke Alexander Wang}\Email{alxwang@cs.stanford.edu}\\
\addr Stanford University
\AND
\Name{Emily B. Fox}\Email{ebfox@stanford.edu}\\
\addr Stanford University \& CZ Biohub
}

\begin{document}

\maketitle

\begin{abstract}
Diabetes encompasses a complex landscape of glycemic control that varies widely among individuals.
However, current methods do not faithfully capture this variability at the meal level.
On the one hand, expert-crafted features lack the flexibility of data-driven methods; on the other hand, learned representations tend to be uninterpretable which hampers clinical adoption.
In this paper, we propose a hybrid variational autoencoder to learn interpretable representations of CGM and meal data.
Our method grounds the latent space to the inputs of a mechanistic differential equation, producing embeddings that reflect physiological quantities, such as insulin sensitivity, glucose effectiveness, and basal glucose levels.
Moreover, we introduce a novel method to infer the glucose appearance rate, making the mechanistic model robust to unreliable meal logs.
On a dataset of CGM and self-reported meals from individuals with type-2 diabetes and pre-diabetes, our \emph{unsupervised} representation discovers a separation between individuals proportional to their disease severity.
Our embeddings produce clusters that are up to 4x better than naive, expert, black-box, and pure mechanistic features.
Our method provides a nuanced, yet interpretable, embedding space to compare glycemic control within and across individuals, directly learnable from in-the-wild data.
\end{abstract}

\section{Introduction}
\begin{table*}[t]
\resizebox{\linewidth}{!}{%
\begin{tabular}{ccccc}
\multicolumn{1}{l}{}                                                    & \begin{tabular}[c]{@{}c@{}}Expert CGM features\\ (TIR, MAGE, etc.)\end{tabular} & \begin{tabular}[c]{@{}c@{}}Mechanistic\\ ODE features\end{tabular} & \begin{tabular}[c]{@{}c@{}}Pure neural\\ network features\end{tabular} & \begin{tabular}[c]{@{}c@{}}Hybrid VAE\\ features (ours)\end{tabular} \\ \hline
\begin{tabular}[c]{@{}c@{}}Noninvasive,\\ inexpensive\end{tabular}      & \yes                                                                            & \no                                                                 & \yes                                                                    & \textbf{\yes}                                                                  \\ \hline
\begin{tabular}[c]{@{}c@{}}Adaptable\\ to new data\end{tabular}             & \no                                                                             & \no                                                                 & \yes                                                                    & \textbf{\yes}                                                                  \\ \hline
\begin{tabular}[c]{@{}c@{}}Physiologically\\ Interpretable\end{tabular} & \no & \yes                                                                & \no                                                                     & \textbf{\yes}                                                                  \\ \hline
\begin{tabular}[c]{@{}c@{}}Robust to\\ in-the-wild data\end{tabular}     & \yes                                                                            & \no                                                                 & \yes                                                                    & \textbf{\yes}                                                                  \\ \hline
\end{tabular}%
}
\caption{By combining our mechanistic knowledge with the flexibility of neural networks, our hybrid VAE has the best of all worlds compared to existing options.}
\label{tab:comparison}
\end{table*}
Diabetes manifests through a multitude of pathways, resulting in a highly heterogeneous population of individuals grappling with the condition \citep{ha_type_2020}.
Traditional metrics for characterizing diabetes, such as HbA1C levels and time-in-range, offer only a limited perspective \citep{brown_beyond_2019}.
They fall short in capturing the nuanced variations in glycemic control at the meal-level, a critical aspect of daily life that is often targeted as the intervention lever to control disease progression.
In contrast, continuous glucose monitors (CGMs), which sample every few minutes, have opened a new avenue into understanding glycemic control.
The advent of CGMs heralded a host of promising possibilities, from personalized nutrition plans attuned to individual glycemic control \citep{zeevi_personalized_2015,hall_glucotypes_2018,berry_human_2020,ben-yacov_personalized_2021}, to remote monitoring systems that deliver patient care based on their glycemic profiles and disease severity \citep{prahalad_teamwork_2022,ferstad_population-level_2021}.
All of these applications require a fine-grained method to subtype glycemic control and diabetic severity, a task which has proven difficult.

Mechanistic models of glucose dynamics have been considered in the past as tools for characterizing glycemic control \citep{hovorka_pancreatic_1998}. These ordinary differential equations (ODEs) provide \emph{clinically interpretable} axes like insulin sensitivity and glucose effectiveness, which quantitatively summarize the disease severity across patients.
However, these models must be fit for each person individually, requiring an intensive and invasive lab test such as an Oral Glucose Tolerance Test (OGTT), a Mixed Meal Tolerance Test (MMTT), or an Intravenous Glucose Tolerance Tests (IVGTT).
These tests are expensive with long appointment backlogs, and require fasting, intravenous injections or insertions, and hours of immobility in a lab setting.
As a result, these mechanistic measures of glycemic control are largely unused in diabetes diagnosis and management, relegating these mechanistic models to be used only in scientific studies \citep{sugiyama_potential_2022} and bioengineering applications \citep{hovorka_closed-loop_2011}.

Beyond the challenge of acquiring high-quality data to fit such mechanistic models to an individual, the rigid form of these models has furthermore limited their translation to real-world, in-the-wild deployments.  First, these ODE-based models are highly sensitive to inaccuracies in self-reported meal information, such as misestimated carbohydrate content, missing meal timestamps, and misreported meal timestamps, all of which are the norm \citep{landry_adherence_2021}. Likewise, these models cannot readily incorporate additional patient covariates or contextual information --- such as demographics or fine-grained macronutrition data --- requiring a bespoke mechanism for each new physiological pathway \citep{man_uvapadova_2014}.

In this paper, we use physiologically-grounded causal pathways of a mechanistic model to learn an \emph{interpretable} representation of post-meal glycemic responses, while simultaneously addressing the challenges of \emph{in-the-wild data}.
The result is a framework that can characterize glycemic control \emph{using only CGM data and self-reported meal information}, potentially enabling a new low-cost and non-invasive tool for characterizing diabetes severity.

More specifically, we introduce a hybrid variational autoencoder (VAE) that unites a flexible neural network encoder with a causally-grounded mechanistic ODE decoder.
The latent variables act as the inputs to our mechanistic ODE, defining the axes of our clinically-interpretable embedding space.
Simultaneously, our encoder infers an effective glucose appearance rate from CGM and meal logs, making the mechanistic model robust to erroneous meal data.

Our contributions are as follows:
\begin{enumerate}
    \item We introduce a hybrid VAE that incorporates a mechanistic differential equation as its decoder. This allows us to ground the neural network on the underlying physiological processes governing glucose dynamics, pruning the function space to only the mechanistically plausible ones.
    \item We use amortized variational inference to eliminate the need to refit the model for every person-meal pair, overcoming scalability issues of traditional mechanistic models. Our approximate Bayesian inference procedure ensures that the inferred parameters align with our prior knowledge of physiologically plausible parameter ranges.
    \item We show that our hybrid model learns an interpretable mechanistic representation that characterizes glycemic control within and across individuals, resulting in patient stratification that agrees with known symptoms of diabetes pathogenesis, \emph{despite not having access to diagnosis information and labwork data}.
\end{enumerate}
By synergizing the flexibility of deep neural networks with the parsimony of mechanistic models, we produce a novel representation space that aligns with clinical understanding and opens a new avenue for scalable, non-invasive, yet fine-grained profiling of glycemic control\footnote{Our implementation is available at \href{https://github.com/KeAWang/interpretable-cgm-representations}{\color{blue}https://github.com/KeAWang/interpretable-cgm-representations}}.

\section{Background}
\subsection{Mechanistic model of insulin-glucose dynamics.}
The core of our mechanistic model is Bergman's minimal model of glucose regulation \citep{bergman_identification_1979} augmented by a standard two-compartment model of glucose appearance \citep{hovorka_partitioning_2002}
\begin{align}
    \dot{G}(t) &= -X(t)G(t) + S_G(G - G_b) + G_2(t)/\tau_m \label{eqn:ode1}\\ 
    \dot{X}(t) &= -p_2 X(t) + p_2 \cdot S_I I(t)\\
    \dot{G}_1(t) &= -G_1(t)/\tau_m + U_G(t) \label{eqn:rate_of_appearance}\\ 
    \dot{G}_2(t) &= G_1(t)/\tau_m - G_2(t)/\tau_m, \label{eqn:ode4}
\end{align}
where $I$ is the insulin concentration in the body, $G_b$ is the basal glucose level, $X$ is a mediator variable that describes the effect of insulin on glucose, and $U_G(t)$ is the rate of appearance of glucose in the body from meal consumption.
The two-compartment model of glucose appearance smooths out and delays glucose appearance from when food is eaten.
Food enters the system through $U_G(t)$, which is often called the glucose appearance rate.
$U_G$ is often parameterized via fixed functional forms and rescaled according to the amount of carbs that is eaten \citep{wang_learning_2022}.
Accurately learning and estimating $U_G(t)$ for generic meals is an active research area \citep{herrero_simple_2012,eichenlaub_bayesian_2021,wang_learning_2022}.
Here we use the standard assumption that $U_G(t) = u_G(t)/V_G$ where $u_G(t)$ is the rate of carbohydrate intake in mg/min and $V_G$ is the accessible glucose volume. We set $V_G$ to a constant due to its non-identifiability.

For individuals with pre-diabetes or type-2 diabetes who are not on external insulin therapy, $I$ is simply the endogenous insulin $\Iendo$.
In our setting, we do not observe $I$, commonly assumed by prior works that seek to identify these model parameters \citep{kanderian_identification_2009, hovorka_partitioning_2002}, since we wish to apply these models outside of a lab setting.
Hence, we must directly account for the unobserved insulin in our mechanistic model.
We assume the endogenous insulin is proportional to the glucose deviation from its basal levels, a common model for insulin-producing individuals \citep{ruan_modelling_2015}:
\begin{align}
    I(t) = \Iendo(t) = M_I \cdot \max(G(t) - G_b, 0) \label{eqn:ode5}.
\end{align}

\paragraph{Interpreting mechanistic parameters.}
Our method's interpretability comes from the causal grounding behind our model's mechanistic parameters:
\begin{itemize}
    \item $S_G$ is known as glucose effectiveness, governing the body's self-regulating ability to bring glucose levels down to baseline levels without insulin. Diabetes is associated with lower glucose effectiveness.
    \item $M_I$ and $S_I$ are known as insulin \emph{productivity} and insulin \emph{sensitivity}, respectively. Diabetes is associated with lower insulin productivity and sensitivity. 
    \item $1/p_2$ and $\tau_m$ govern physiological timescales, corresponding to the peak effect of insulin and the peak appearance of glucose from the meal, respectively. Diabetes is associated with longer timescales, as it takes longer for blood glucose to be brought down to baseline levels.
\end{itemize}
Together with the basal glucose level $G_b$, these parameters provide a low-dimensional and succinct description of diabetes severity and heterogeneity.
However, estimating these parameters outside of lab settings remains a challenge.

\subsection{Latent Variable Modeling}
In latent variable modeling, we assume there exists a joint distribution $p_\theta(\xbf, \zbf) = p_\theta(\xbf \given \zbf)p(z)$ between our observed data $\xbf$ and a set of latent variables $\zbf$.
Then, for every distribution \(q_\phi(\zbf)\), there exists a variational lower bound on the log evidence/marginal likelihood:
\begin{align}
    \log p_\theta(\xbf) \geq \Lcal(\theta, \phi; \xbf) = \ev[\zbf \sim q_{\phi}]{\log \frac{p_\theta(\xbf, \zbf)}{q_\phi(\zbf)}}.
\end{align}

The evidence lower bound (ELBO) can be rewritten to be amenable to stochastic optimization with reparameterized samples \citep{kingma_auto-encoding_2013,rezende_stochastic_2014,titsias_doubly_2014}:
\begin{align}\label{eqn:elbo}
\begin{split}
\Lcal(\theta, \phi; \xbf) &= \ev[\zbf \sim q_\phi]{\log p_\theta(\xbf \given \zbf)} \\
&\quad - \beta \cdot \kldiv{q_\phi(\zbf)}{p(\zbf)}
\end{split}
\end{align}
for $\beta \geq 1$.
The ELBO decomposes into a sum of two terms, a reconstruction term and a regularization term.
Tuning $\beta$ trades off accurate reconstruction and sample quality for disentangled representations \citep{bowman_generating_2016,higgins_beta-vae_2016}.

The decoder $p_\theta$ and encoder $q_\phi$ are usually parameterized by neural networks with parameters $\theta$ and $\phi$ that are trained by maximizing \autoref{eqn:elbo} with reparameterized sampling.
Solving this optimization problem by learning these parametric functions is known as amortized variational inference.
The amortization comes from reusing these learned parametric functions to do posterior inference on new observations, instead of having to solve an optimization problem every time \citep{blei_variational_2017},

In this work, we instead parameterize the decoder with our mechanistic model, providing us with a low-dimensional and interpretable latent space. %

\section{Mechanistic Hybrid Variational Autoencoder}
Our goal is to embed high-dimensional observations of glycemic response in a clinically-interpretable latent space.  Such representations can be used for downstream tasks such as studying patient heterogeneity and subtypes, characterizing disease progression, or designing personalized nutrition.  More formally, let $\xbf=(x_1, \ldots, x_T)$ denote a sequence of observations, such as blood sugar measurements from a continuous glucose monitor (CGM).
Assume that every observation is associated with a context variable $\cbf$, such as demographics or a sequence of meals an individual ate at the corresponding timesteps $(c_1, \ldots, c_T)$.
We want to infer a latent variable $\zbf$ associated with each $\xbf$.  To provide clinical interpretability, we propose that $\zbf$ (or a transform thereof) correspond to the parameters of our mechanistic model.

Notice that we can freely replace the distribution $p_\theta(\xbf \given \zbf)$ and $q_\phi(\zbf)$ in \autoref{eqn:elbo} with conditional distributions, resulting in the following conditional ELBO:
\begin{align}\label{eqn:celbo}
\begin{split}
\Lcal(\theta, \phi; \xbf, \cbf) &= \ev[\zbf \sim q_\phi(\zbf \given \cbf)]{\log p_\theta(\xbf \given \zbf, \cbf)} \\
&\quad - \beta \cdot \kldiv{q_\phi(\zbf\given \cbf)}{p(\zbf \given \cbf)}.
\end{split}
\end{align}
To reduce the number of trainable parameters in our model for our data-scarce application, we make the simplifying assumption that $p(\zbf \given \cbf) = p(\zbf)$, though past works have found the conditional prior beneficial in other applications \citep{sohn_learning_2015}.
In our experiments, we found that it was very important to initialize the prior to a fixed expert prior based on what is physiologically plausible.
See \autoref{app:initialization} for more details.

\paragraph{Grounding the latent space}
So far, we have kept the parameterization of our VAE completely general.
Indeed, typical parameterizations use a neural network for both the encoder $q_\phi$ and the decoder $p_\theta$ to provide enough flexibility to approximate the posterior distribution and generate accurate reconstructions.
Unfortunately, a purely black-box parameterization will produce an uninterpretable latent space.
For our health application, we would like interpretable representations that also make use of our existing causal understanding of the body.

To ground the latent space in physiologically meaningful quantities, we return to the mechanistic ODEs defined by \autoref{eqn:ode1} through \autoref{eqn:ode5}.
\emph{We parameterize the decoder with the Bergman model with endogenous insulin}, using a sequence of discrete Euler integration steps to autoregressively model $p_\theta$.
That is, we have
\(
    p_\theta(\xbf\given \zbf, \cbf) = p_\theta(x_1\given \zbf, \cbf) \prod_{i=2}^{T} p_\theta(x_i \given \zbf, x_{i-1}, \cbf),
\)
where each factor is a normal distribution centered at an integration step output.
To retain sufficient flexibility, we parameterize the encoder \(q_\phi\) with a factorized normal distribution with means and scales output by an LSTM.

Our parameterization constrains the latent space to be the inputs of our mechanistic ODE. These inputs include the ODE initial state $\xbf_0=[G(0), X(0), G_1(0), G_2(0)]$ and the ODE parameters $\wbf=[\tau_m, G_b, S_G, p_2, S_I, M_I]$.

\paragraph{Inferring effective glucose appearance}
Recall that a key component of the Bergman minimal model is the rate of appearance of glucose $U_G(t)=u_G(t)/V_G$ from \autoref{eqn:rate_of_appearance}.
However, in our in-the-wild setting the carbohydrate consumption rate $u_G$ is known only through self-reported data, which frequently contain inaccurate meals or misreported timestamps.
Such errors are especially challenging for mechanistic ODEs; for example, meals that are logged after the glucose rises violate the causal structure of the ODE.

\begin{figure}[H]
    \centering
    \includegraphics[width=\linewidth]{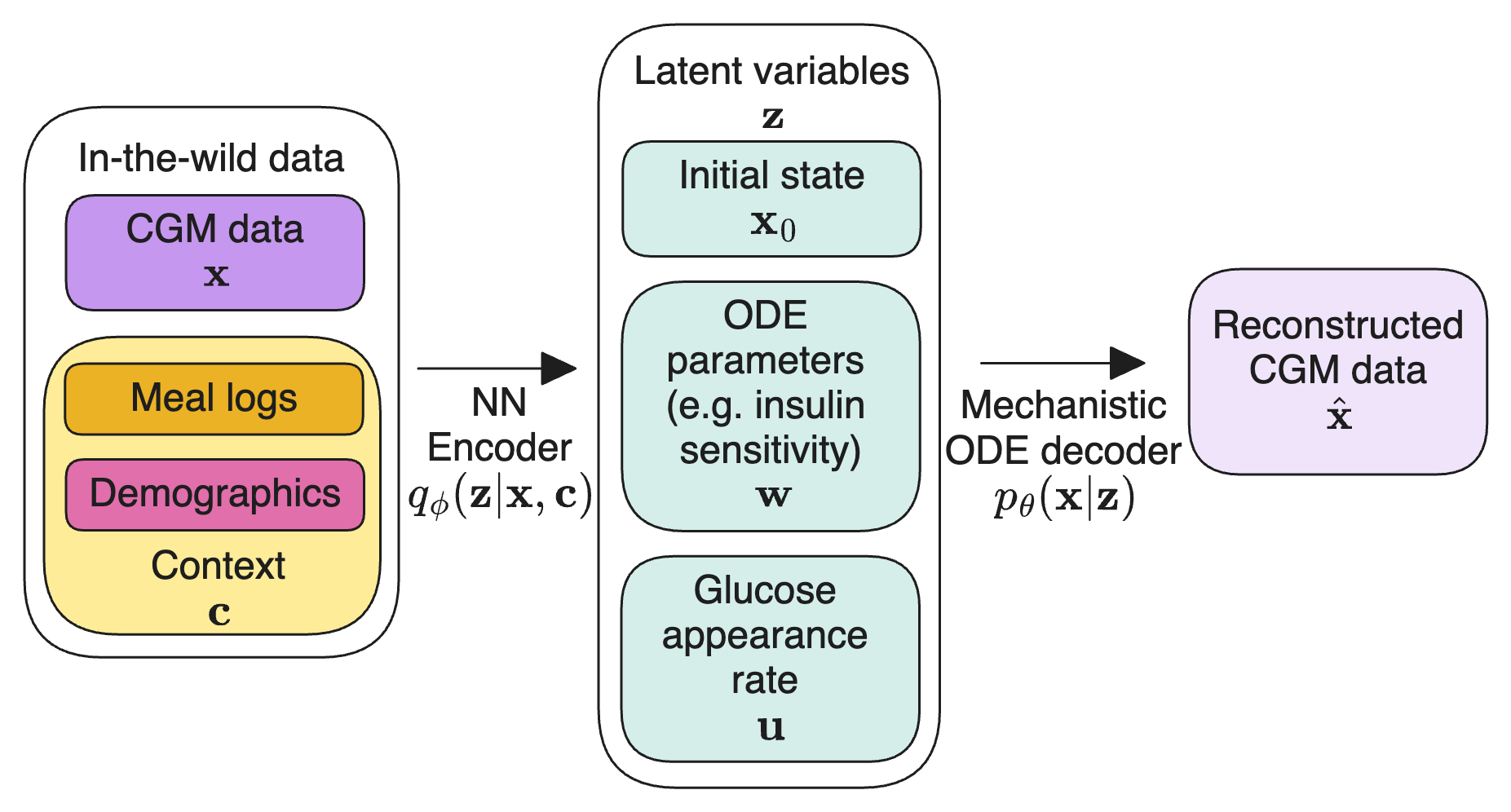}
    \caption{Visualization of our mechanistic hybrid VAE. Notably, our unsupervised method can be applied directly to in-the-wild data.}
    \label{fig:model_figure}
\end{figure}

To make our model robust to data issues with meal information, we propose to use the encoder to infer an ``effective'' carb consumption $\ubf$ from data, by including it as a latent variable.
Here $\ubf=(u_1,\ldots, u_T)$ where $u_i=u_G(t_i)$.
This is possible because rises and falls in CGM data contain information about when a meal was eaten and how much was eaten.
By inferring $\ubf$, our encoder learns to extract this information from CGM, assisted by the meal information, without being restricted to the reported carb consumption.
See \autoref{fig:model_figure} for a visualization of this mechanism.

\paragraph{Practical implementation}
We would like our latent variable $\zbf$ to represent three sets of variables: $\ubf$, $\xbf_0$, and $\wbf$.
However, all of these variables must be constrained to be non-negative and fall in physiologically plausible ranges, which is inconvenient to directly model probabilistically.
Instead, we model these parameters in an unconstrained space.
Thus we define $\ubf = \sigma_\ubf(\ubf')$, $\xbf_0 = \sigma_{\xbf_0}(\xbf_0')$, and $\wbf = \sigma_\wbf(\wbf')$ where each $\sigma_{(\cdot)}$ is an invertible transformation (e.g. rescaled sigmoid) that maps $\reals$ to a constrained space $[a_{(\cdot)}, b_{(\cdot)}]$.
Thus, we finally arrive at our latent variable $\zbf=(\ubf', \xbf_0', \wbf')$, which allows us to do variational inference over an unconstrained space.
When decoding, we map the unconstrained latent variables into a physiologically-plausible constrained space for both the parameters and the initial states.
See \autoref{app:constaints} for details on the constraint ranges.

To account for missing values in $\xbf$ and $\cbf$, we project each timestep into separate 8-dimensional embedding spaces and mask the entire timestep in embedding space when the corresponding timestep has a missing channel as done in \citep{yue_ts2vec_2021}.

\section{Dataset}\label{sec:dataset}
In our experiments we use the Keto-Med dataset \citep{landry_adherence_2021}, a randomized crossover trial (NCT03810378) that placed participants on two different low-carb diets, Ketogenic and Mediterranean.
The participants were given controlled meals during the first four out of twelve weeks for each diet, but were responsible for their own diet adherence for the last eight weeks.
The study targeted adults from the San Francisco Bay area with Type-2 diabetes
or prediabetes.
Only a subset of recorded meals have accompanying glucose measurements, since participants were only given CGMs periodically during the study.
The meal information in this dataset is representative of data-collection challenges for in-the-wild conditions, including missing meals, inaccurate meal information, and mistimed reporting.
From each meal, we extract the total amount of food eaten, total carb, total sugar, total dietary fiber, total fat, and total protein as the meal covariates.
Note that not all meals contain all of these macronutritional information.

After preprocessing and joining the data, we end up with a total of 964 postprandial glucose response (PPGR) time-series from 33 individuals.
We define a PPGR to be a sequence of length $T=60$ (5 hours), containing CGM and self-reported information, with a food intake occurring at timestep $t=12$ (hour 1).
We include 1 hour of context before the meal to give the model information about glucose excursions prior to the meal.
The 4-hour horizon following the meal is set to be sufficiently long to capture the temporal effects of the meal.
Note that often other foods are eaten during this 5 hour period, making the problem of estimating mechanistic parameters even more difficult.

\section{Results}\label{sec:results}
We apply our hybrid VAE model to the CGM and meal dataset described in \autoref{sec:dataset}.
We compare to a baseline non-hybrid VAE model with a similar number of parameters (53,561) as our model (41,454).
Both models use a 2-layer bidirectional LSTM encoder with a hidden state size of 32.
The non-hybrid VAE uses a 1-layer uni-directional LSTM decoder with the same hidden state size as well, as a 32-dimensional black-box latent space $\zbf$.
Both models use $\hat{\beta}=0.01$ as defined in \citet{higgins_beta-vae_2016}, a batch size of 64, and the ADAM \citep{kingma_adam_2015} optimizer with a constant learning rate of $0.01$. Our model converged quickly in 100 epochs of training, while the black-box model was trained for 1000 epochs.
Importantly, we use a high dropout rate of 0.5 following the latent $\ubf$ to regularize the inferred $\ubf$ away from non-plausible meal sequences \citep{srivastava_dropout_2014}.
We tune parameters for both models to train until convergence and achieve similar reconstruction losses. 
Both models are trained on CGM data, meal data, and demographics data (age, weight, and sex).
Static variables like demographics are repeated and concatenated with the per-timestep meal data as $\cbf$.
We leave out additional contextual variables such as BMI and HbA1C from model training since these variables are highly correlated with the hidden diagnosis information, potentially biasing the models' representations.

\begin{figure}[H]
    \centering
    \includegraphics[width=\linewidth]{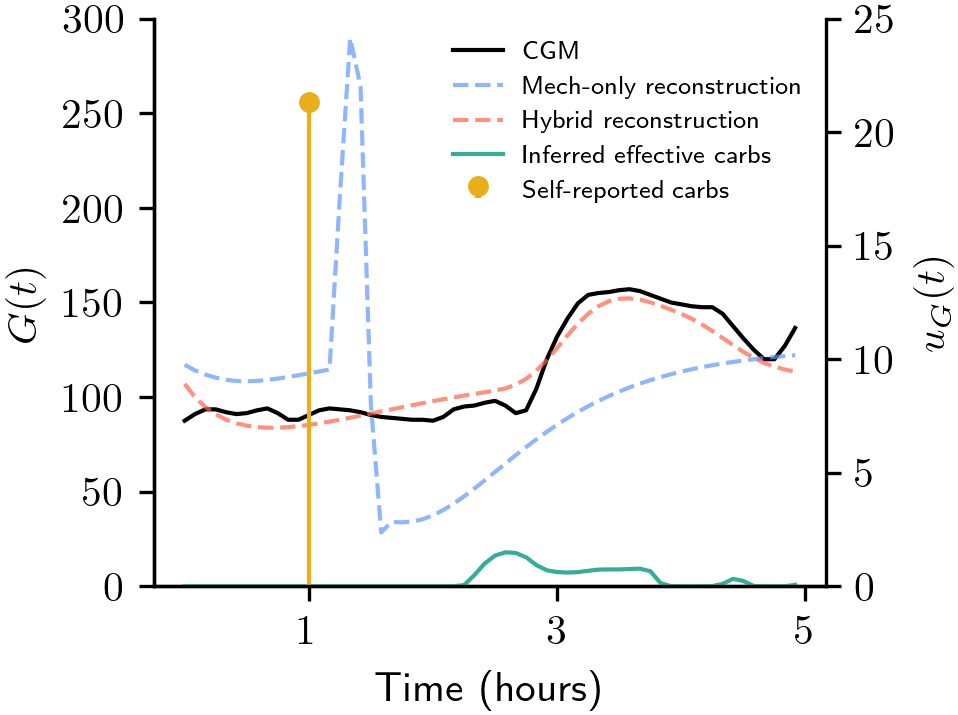}
    \caption{The mechanistic-only model is brittle to incorrect self-reported meal data. Here, it is unable to explain the delayed rise in glucose due to the early reporting. However, our hybrid model can compensate for this and infer an effective carb rate, leading to a more accurate reconstruction.}
    \label{fig:mech_vs_hybrid}
\end{figure}
\begin{figure*}[h!]
    \centering
    \includegraphics[width=\linewidth]{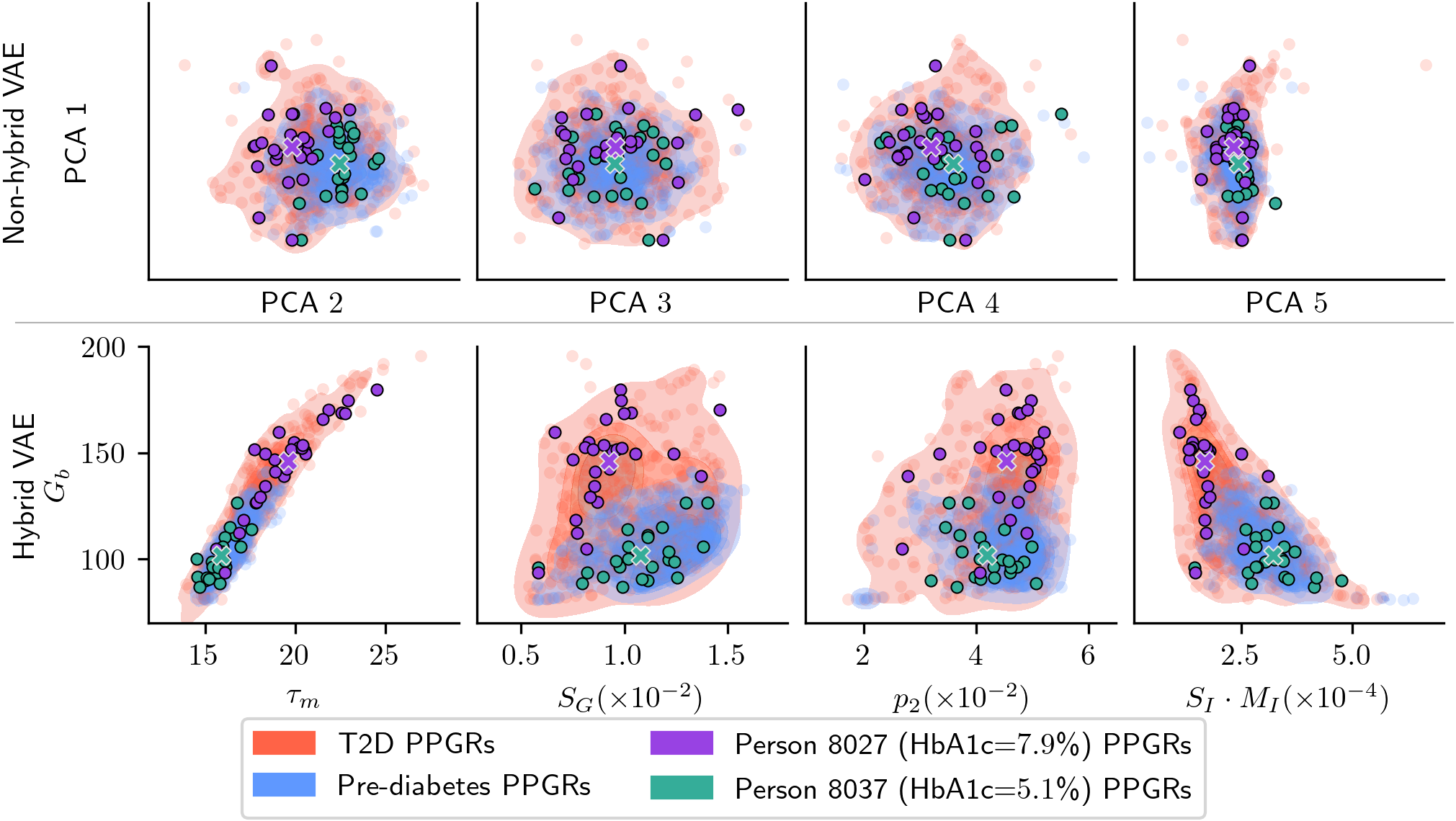}
    \caption{Post-prandial glucose responses (PPGRs) for all meals projected into a non-hybrid VAE's embedding space (top) and our mechanistic embedding space (bottom). Our embeddings are learned \emph{without} any diagnosis information or diabetes indicators besides CGM. The embeddings of the two groups overlap as expected. Notably, the distinction between the two groups aligns with our understanding of diabetes disease progression; individuals with diabetes have higher basal glucose $G_b$, more extended meal responses $\tau_m$, and lower insulin sensitivity and insulin productivity ($S_I \cdot M_I$). We also show the average embedding for two individuals on opposite ends of the disease spectrum (8027 and 8037), outlined in white.}
    \label{fig:offset_vs_params}
\end{figure*}
\begin{figure*}[h!]
    \centering
    \includegraphics[width=\linewidth]{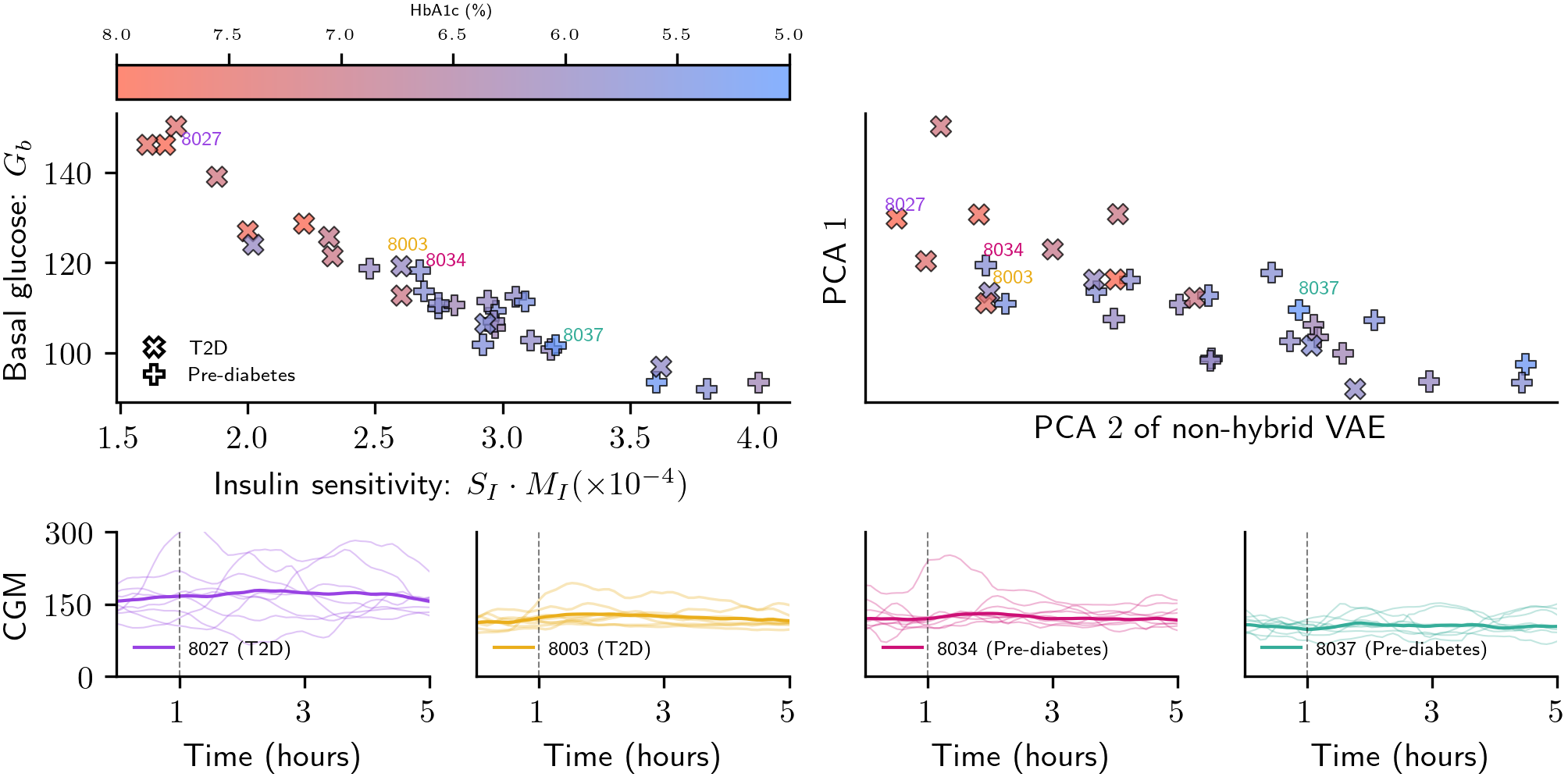}
    \caption{Here we show the average of each patient's PPGR embeddings in both representations. Notice that the mechanistic embeddings are not only more interpretable but also more consistent with HbA1C characterizations of disease progression.
    Patients that are similar in HbA1C tend to also be closer together in our mechanistic space, even though the model was not trained on HbA1C.
    Not only are our representations more interpretable, they also provide better separation.
    A linear decision boundary on this simple 2D plane can correctly classify 29 out of 33 individuals versus 26 on the black-box PCA plane.
    We choose a subset of individuals to visualize the disease heterogeneity as characterized by CGM data along this spectrum.
    }
    \label{fig:avg_params}
\end{figure*}

\paragraph{Robustness to self-reported meals}
We first show the robustness of our model to mistimed meal reports in \autoref{fig:mech_vs_hybrid}.
We train the mechanistic model by doing direct gradient descent on its parameters and initial state with respect to mean-squared error.
The reported meal sequence here is typical of a standard PPGR in-the-wild.
Although macronutritional components, such as high fat content, can delay the rise of glucose \citep{ann_ahern_exaggerated_1993}, \autoref{fig:mech_vs_hybrid} is a clear case of a meal that was logged \emph{too early} with a possible carb overestimate.
As a result, the brittle mechanistic-only model must contort the mechanistic parameters to describe the observed CGM data, leading to poor reconstructions.
In contrast, our hybrid VAE infers an ``effective'' meal consumption rate $u_G$ that can explain the data much better.
Although not shown here, the mechanistic-only model also fails to account for rises in glucose when carb information is missing entirely or reported late.
Late and missing meals logs are especially challenging for mechanistic-only models, since they violate the causality of the ODE.

\paragraph{Physiologically accurate meal-level embedding space}
\autoref{fig:offset_vs_params} shows the distribution of all meal responses (PPGRs) embedded into the latent space of both respective models.
Since we do not observe the true insulin of the individuals, $S_I$ and $M_I$ are not identifiable on their own, only their product is identifiable. Hence we visualize $S_I \cdot M_I$ in our results.

By encoding the data with physiologically meaningful variables, our hybrid VAE can naturally separate individuals with pre-diabetes from individuals with type-2 diabetes \emph{despite not being given any diagnosis information}.
Moreover, the two populations are characterized in a way that is directly in line with our clinical understanding of the progression from pre-diabetes to Type-2 diabetes. Through its mechanistic grounding, the hybrid VAE can discover not just higher basal glucose levels $G_b$ and longer elevated glucose levels $\tau_m$ among people with diabetes, \emph{but also reduced glucose effectiveness $S_G$ and insulin sensitivity $S_I\cdot M_I$}.
In addition, we see a clear separation \emph{in every dimension} between the average embedding for person 8037 and 8027, who have the lowest and highest levels of HbA1C in the dataset, respectively.
Our results show that our meal-level mechanistic embedding space can serve as a powerful way to visualize \emph{intrapatient} variations in glycemic control.

In contrast, 2D principal component analysis (PCA) projections of the non-hybrid model's latent space show 
a higher overlap between the two populations. In fact, in all but one of the projections (PCA 1 vs PCA 2), the PPGRs from both populations are completely interspersed. Although there is a slight separation in the PCA 1 vs PCA 2 projection, it is unclear what the two axes PCA 1 and PCA 2 correspond to physiologically.

\paragraph{Interpretable characterization of disease progression}
In addition to intrapatient variation, our mechanistic representation can effectively characterize \emph{interpatient} variation in disease progression.
\autoref{fig:avg_params} (top) shows the average embedding by both models for every individual in the dataset along with their HbA1C levels, the current gold standard of characterizing diabetes severity.
Note that HbA1C was not used to train either model.

From this visualization, we see that our hybrid-VAE's learned basal glucose $G_b$ and insulin sensitivity $S_I\cdot M_I$ correlates well with increasing HbA1C levels. Although it is known that HbA1C can be well estimated by baseline glucose levels, here we also show its correlation with reduced insulin sensitivity.

Although the black-box embeddings of the non-hybrid VAE model shows a similar trend along its first two principal components, the model wrongly places a person with diabetes with one of the highest HbA1C levels in the study into a cluster of medium-level HbA1C individuals.

\paragraph{Evaluating Embedding Quality}
In \autoref{tab:clustering-scores} we quantitatively evaluate our Hybrid VAE's embedding space relative to 6 other baselines, separated into ``naive'', ``expert'', ``black-box'', and ``graybox'' categories. 
For each method, we represent an individual by averaging their embeddings across all their PPGRs.
We then perform $k$-means clustering with $k=2$.
We evaluate each method by standard clustering metrics: normalized mutual information (NMI), adjusted mutual information (AMI), homogeneity, and completeness, as implemented in \citep{pedregosa_scikit-learn_2011}.
All metrics fall in the range of $[0,1]$ with 1 being the highest.
We use the diagnosis: ``pre-diabetes'' vs ``type 2 diabetes as ground truth clusters.
Our Hybrid VAE consistently produces the highest quality low-dimensional embedding space, with significantly better interpretability.

The naive baselines include ``Raw CGM'' and ``Raw CGM + DTW''. For these two baselines, we use the original raw CGM space as the embedding space.
The latter uses $k$-means with Dynamic Time Warping (DTW), a standard time-series clustering method \citep{sakoe_dynamic_1978}, as implemented by \citet{tavenard_tslearn_2020}.
Note that ``CGM'' and ``CGM + DTW'' do not involve a low-dimensional embedding space: they operate in the original raw sensor space.

The expert features include a set of 10 well-established CGM-derived features that are used to clinically measure glycemic variability and control \citep{kovatchev_metrics_2017}.
The features are: mean glucose, glucose standard deviation, glucose coefficient of variability (CV) \citep{bergenstal_recommendations_2013}, max glucose value, min glucose value, percent time in range (TIR) as recommended by the American Diabetes Association guidelines \citep{american_diabetes_association_professional_practice_committee_6_2021}, glucose arc length \citep{peyser_glycemic_2018}, J-index \citep{wojcicki_j-index_1995}, mean amplitude of glucose excursion (MAGE)\citep{service_mean_1970}, incremental area under the glucose curve relative to the starting baseline (iAUC) \citep{brouns_glycaemic_2005}. 
We excluded other metrics such as eA1C \citep{bergenstal_glucose_2018}, CONGA \citep{mcdonnell_novel_2005}, and MODD \citep{molnar_day--day_1972}, as they are long-term glycemic variablility metrics and require longer CGM sequences than our 6 hour PPGR period. Nevertheless, these excluded metrics are closely related to the metrics we chose; for example, eA1C is merely an affine transform of the mean glucose.

Besides comparing to the latent space of the black-box VAE described in \autoref{sec:results}, we also compare to Time Contrastive Learning (TCL) \citep{hyvarinen_unsupervised_2016}, a strong baseline for learning black-box embeddings.
We evenly divide each PPGR into 3 time-windows for TCL, using a 2 hidden-layer MLP with 32 hidden dimensions, corresponding to 5,667 trainable parameters.
The output of the final pre-activation layer is used as the embedding space.
Since TCL produces an embedding for each time-window, we compare averaging the time-window embeddings against concatenating the time-window embeddings.

Finally, we compare to ``Mechanistic ODE'', which uses the ODE parameters of a non-neural mechanistic model, i.e. the decoder of our hybrid ODE, as its embedding space.
\autoref{tab:clustering-scores} confirms the stark contrast hinted by \autoref{fig:mech_vs_hybrid}. Hybridizing the mechanistic ODE results in 3 to 4 times higher clustering scores, showing the need to augment the original mechanistic model with more flexible components.

\begin{table}[h]
\resizebox{\linewidth}{!}{%
\begin{tabular}{lrrrr}
\hline
                              & \multicolumn{1}{l}{NMI} & \multicolumn{1}{l}{AMI} & \multicolumn{1}{l}{Hom.} & \multicolumn{1}{l}{Comp.} \\ \hline
Raw CGM $(d=60)$                  & 0.41                    & 0.39                    & 0.36                            & 0.47                             \\
Raw CGM + DTW $(d=60)$            & {\ul 0.54}              & {\ul 0.53}              & \textbf{0.62}                   & {\ul 0.51}                       \\ \hline
Expert features ($d=10$)   & 0.25               & 0.25              & {\ul{0.61}}                   & 0.21                       \\ \hline
TCL + Average $(d=32)$        & 0.24                    & 0.21                    & 0.19                            & 0.33                             \\
TCL + Concat $(d=3\times 32)$ & 0.29                    & 0.27                    & 0.24                            & 0.38                             \\
Black-box VAE $(d=32)$                  & 0.21                    & 0.19                    & 0.22                            & 0.21                             \\ \hline
Mechanistic ODE $(d=7)$       & 0.14                    & 0.12                    & 0.14                            & 0.14                             \\
Hybrid VAE $(d=7)$            & \textbf{0.54}           & \textbf{0.53}           & 0.51                      & \textbf{0.58}                    \\ \hline                   
\end{tabular}%
}
\caption{Embedding quality as measured by $k$-means clustering on each individual's average embedding, using normalized mutual information, adjusted normalized mutual information, homogeneity, and completeness. Higher is better. ``Pre-diabetes'' and ``type 2 diabetes'' labels form the ground truth clusters. Our embeddings produce up to 4x better clusterings than other methods.}
\label{tab:clustering-scores}
\end{table}

\section{Related Work}
Until recently, mechanistic models have been primarily used for system identification in systems with few degrees of freedom \citep{hodgkin_quantitative_1952,varah_spline_1982,bergman_identification_1979,leander_stochastic_2014}.
Many recent works have explored ways to combine the inductive biases of mechanistic knowledge with the flexibility of black-box machine learning models. These models are also known as hybrid models \citep{miller_breimans_2021,levine_framework_2021} or gray-box models \citep{duun-henriksen_model_2013}, and they are especially relevant in healthcare and biological settings where data efficiency and interpretability are paramount \citep{baker_mechanistic_2018}.

In healthcare applications, recent works have shown that hybrid models can improve forecasting and prediction performance compared to black-box models when data is scarce. \citet{miller_learning_2020} used a highly flexible mechanistic model for Type-1 diabetes to produce more physiologically plausible multi-hour CGM forecasts compared to purely black-box models and classic forecasting techniques.
\citet{hussain_neural_2021} used an attention-based hybrid model that ensembles multiple mechanistic pathways from pharmacological modeling, allowing them to more accurately model cancer disease progression. In parallel, \citet{qian_integrating_2021} augments a mechanistic state space pharmacokinetics-pharmacodynamics model with black-box latent variables to model patient biomarkers longitudinally.

Outside of healthcare, hybrid models have been successfully applied in engineering and physics applications where the mechanistic models are better specified \citep{takeishi_physics-integrated_2021, yin_augmenting_2021,wehenkel_robust_2022,karniadakis_physics-informed_2021}.

There have also been prior works that seek to predict post-prandial glycemic responses \citep{ben-yacov_personalized_2021}, characterize diabetes disease progression \citep{wilson_cgm_2023}, and subtype glycemic control \citep{keshet_cgmap_2023}.
However, these works rely solely on expert hand-crafted features, which are lower quality than our learned embeddings, as demonstrated in \autoref{tab:clustering-scores}.

In contrast to prior literature that primarily focus on predictive accuracy or disease subtyping with handcrafted features, our work \emph{learns} an interpretable representation that is useful for comparing disease variation using \emph{in-the-wild data}, \emph{without needing label information}.

\section{Discussion}
The advent of continuous glucose monitors has opened a new frontier in understanding individualized glycemic control, offering a level of granularity previously unattainable. Yet, the full potential of this rich data source remains untapped due to the limitations of existing analytical tools. Our work addresses a part of this gap by marrying the flexibility of neural networks with the causal structure inherent in metabolic mechanistic models.
Our hybrid approach enables us to learn physiologically grounded representations directly from in-the-wild data, without the constraints of invasive lab tests.
Indeed, we can have \emph{both} interpretability and separability in a single embedding space, all learned from out-patient data.

Traditional mechanistic models, while grounded in physiological understanding, often falter when applied to self-reported data. They struggle with missing meals, inaccurate meal estimates, and other imperfections that are common in outpatient settings. In contrast, our method is designed with these challenges in mind while remaining scalable and practical for real-world data.

Our hybrid VAE has shown to be a powerful tool for subtyping interindividual and intraindividual glycemic control, which is a key step in applications like personalized nutrition \citep{ben-yacov_personalized_2021} and remote health monitoring.
As CGMs become ever more accessible, our framework has the promise of becoming a new low-cost component for the next generation of non-invasive patient-facing diabetes care.

\reveal{
\acks{This work was supported in part by AFOSR Grant FA9550-21-1-0397, ONR Grant N00014-22-1-2110, National Science Foundation Grant 2205084, the Stanford Institute for Human-Centered Artificial Intelligence (HAI). KAW was partially supported by Stanford Data Science as a Stanford Data Science Scholar. EBF is a Chan Zuckerberg Biohub – San Francisco Investigator. 

We thank Matthew J. Landry and Christopher Gardner for providing us access to the Keto-Med dataset. We thank David Maahs, Priya Prahalad, Jiaxin Shi, and Rahul G. Krishnan for insightful discussions.

We thank the creators and maintainers of open-source software we relied on that made this work possible \citep{pedregosa_scikit-learn_2011,virtanen_scipy_2020,van_rossum_python_1995,harris_array_2020,paszke_pytorch_2019,tavenard_tslearn_2020,hunter_matplotlib_2007}.
}
}

\bibliography{wang23}

\appendix
\section{Initializing and fixing the expert prior}\label{app:initialization}
We found that it was important to initialize our prior distribution $p(\zbf)$ to physiologically plausible values and kept fix during training.
This is consistent with the traditional approach of specifying an expert-informed Bayesian prior over the latent variables.

For the prior over $\ubf'=\sigma_{\ubf}^{-1}(\ubf)$, we set the mean to 0 and standard deviation to $10$ for each timestep.
For the prior over $\xbf_0'=\sigma_{\xbf_0}^{-1}(\xbf_0)$ where $\xbf_0=[G(0), X(0), G_1(0), G_2(0)]$, we set the mean to be result of mapping $[30, 120, 10^{-2}, 5\times 10^{-4}, 1/30, 1]$ to the unconstrained space; we set the standard deviation to be the result of mapping $[2,1,1,1,1,1]$ to the unconstrained space.
For the prior over $\wbf'=\sigma_{\wbf}^{-1}(\wbf)$ where $\wbf=[\tau_m, G_b, S_G, p_2, S_I, M_I]$, we set the mean to the result of mapping $[120, 0.1, 0.1, 20]$ to the unconstrained space and the standard deviatino to the result of mapping $[1,1,1,2]$ to the unconstrained space.

\section{Parameter and state constraints}\label{app:constaints}
Since our hybrid VAE relies on a mechanistic ODE decoder, the initial state and the parameters of passed into the decoder must be physiologically-plausible.
These constraints regularize the mechanistic ODE to be more numerically stable, more identifiable, and ensure that the learned latent space is physiologically interpretable and comparable to real values.

We use the following constraints on the initial state $\xbf_0=[G(0), X(0), G_1(0), G_2(0)]$:
\begin{itemize}
    \item $G(0)\in [50, 300]$ mg/dL
    \item $X(0)\in [0, 1]$ min$^{-1}$
    \item $G_1(0) \in [0, 1]$ mg/dL
    \item $G_2(0) \in [0, 100]$ mg/dL.
\end{itemize}
We use the following constraints on each ODE parameter $\wbf=[\tau_m, G_b, S_G, p_2, S_I, M_I]$:
\begin{itemize}
    \item $\tau_m \in [10, 60]$ min
    \item $G_b \in [80, 200]$ mg/dL
    \item $S_G \in [5\times 10^{-3}, 2 \times 10^{-2}]$ min$^{-1}$
    \item $S_I \in [10^{-4}, 10^{-3}]$ (L/mU)/min 
    \item $p_2 \in [1/60, 1/15]$ min$^{-1}$
    \item $M_I \in [0.1, 3.0]$ (mU/L)/(mg/dL).
\end{itemize}
We constrain the carb amount $u_t$ to be $[0, 1000]$ mg/min.

\end{document}